\documentclass{article}

\usepackage{arxiv}

\usepackage{url,hyperref,lineno,microtype,subcaption}
\usepackage[onehalfspacing]{setspace}
\usepackage{listings}
\usepackage{graphicx}

\title{Nengo and low-power AI hardware for robust, embedded neurorobotics} 

\author{
    Travis DeWolf\\
	Applied Brain Research\\
	Waterloo, ON \\
	\And
	Pawel Jaworski\\
	Applied Brain Research\\
	Waterloo, ON\\
	\And
	Chris Eliasmith\\
	Applied Brain Research\\
	University of Waterloo\\
	Waterloo, ON\\
}

\begin{document}
%\onecolumn
%\firstpage{1}

\maketitle

\begin{abstract}
In this paper we demonstrate how the Nengo neural modeling and simulation libraries enable users to quickly develop robotic perception and action neural networks for simulation on neuromorphic hardware using tools they are already familiar with, such as Keras and Python.
We identify four primary challenges in building robust, embedded neurorobotic systems, including: 1) developing infrastructure for interfacing with the environment and sensors; 2) processing task specific sensory signals; 3) generating robust, explainable control signals; and 4) compiling neural networks to run on target hardware.
Nengo helps to address these challenges by: 1) providing the NengoInterfaces library, which defines a simple but powerful API for users to interact with simulations and hardware; 2) providing the NengoDL library, which lets users use the Keras and TensorFlow API to develop Nengo models; 3) implementing the Neural Engineering Framework, which provides white-box methods for implementing known functions and circuits; and 4) providing multiple backend libraries, such as NengoLoihi, that enable users to compile the same model to different hardware.
We present two examples using Nengo to develop neural networks that run on CPUs and GPUs as well as Intel's neuromorphic chip, Loihi, to demonstrate two variations on this workflow.
The first example is an implementation of an end-to-end spiking neural network in Nengo that controls a rover simulated in Mujoco.
The network integrates a deep convolutional network that processes visual input from cameras mounted on the rover to track a target, and a control system implementing steering and drive functions in connection weights to guide the rover to the target.
The second example uses Nengo as a smaller component in a system that has addressed some but not all of those challenges. Specifically it is used to augment a force-based operational space controller with neural adaptive control to improve performance during a reaching task using a real-world Kinova Jaco$^2$ robotic arm.
The code and implementation details are provided~\footnote{\url{https://github.com/abr/neurorobotics-2020}}, with the intent of enabling other researchers to build and run their own neurorobotic systems.
\tiny
\keywords{ Nengo, neuromorphic, neurorobotic, spiking neural networks, robotic control, adaptive control, embedded robotics}
\end{abstract}

\section{Introduction}
Specialized AI hardware offers exciting potential for the development of low-power, highly responsive robotic systems with embedded control.
Edge devices for accelerating neural networks are starting to become commercially available from companies such as NVIDIA, BrainChip, GrAI Matter Labs, Google, Intel, and IBM.
While the promise of low-power and low-latency embedded processing and control is highly desirable, the process of implementing algorithms on the hardware generally remains a significant hurdle.
Developing neural networks for processing sensory data and generating control signals is a difficult problem, and adding further constraints specific to a particular piece of hardware only increases the challenge.
In this paper we focus on the development of spiking neural networks (SNNs) for the subset of devices known as neuromorphic hardware \cite{mead1990neuromorphic}. Effectively using such hardware often requires additional expert knowledge outside of traditional machine learning and neural network methods to program effectively.
In short, it is difficult to quickly and easily build robust, integrated neural models for controlling robots using neuromorphic hardware.

Building neurorobotic systems can be characterized as consisting of four tasks: 
\begin{enumerate}
    \item Developing infrastructure to send and receive signals from the environment.
    There are a multitude of different interface protocols for sensors, hardware, and simulators.
    To minimize development time, simple interfaces should be available and interchangeable with minimal changes to the model description.
    \item Processing task specific sensory signals.
    Deep neural networks (DNNs) are the principle machine learning tool used for sensory processing, and it is important to take advantage of the extensive literature and solutions in this field.
    To that end, users need to be able to take DNNs, convert them to networks that can run on neuromorphic hardware, and integrate them into a neurorobotics control system.
    For systems using perception methods not rooted in neural networks, it is also important to be able to easily integrate their output with downstream networks.
    \item Generating robust control signals with explainable neural networks.
    When generating control signals, having guarantees on performance is important, and often necessary.
    To accomplish this users needs to know exactly what operations are being implemented to guarantee stability.
    The Neural Engineering Framework (NEF; \cite{eliasmith2004neural}) offers `white-box' neural network development methods that allow integration of these methods into neurorobotics control systems, making an API for building up such networks quickly desirable.
   \item Compiling neural networks to run on multiple targeted hardware platforms.
    During the process of designing control and perception systems it is often desirable to develop neural network models on standard hardware with minimal compilation overhead.
    Once a prototype network is working, it should be straightforward to compile to targeted special purpose hardware.
    Being able to compile the same model to different hardware can greatly speed up the development of neurorobotics systems.
\end{enumerate}

In this paper we present a neurorobotics development workflow for building neural networks that run on standard and neuromorphic hardware using the Nengo neural modeling platform (\url{http://nengo.ai/}; \cite{bekolay2014nengo}).
As part of this workflow, we take advantage of the NengoInterface package to streamline interfacing with the physics simulators, the NengoDL package for integrating Keras and TensorFlow models that process incoming sensory data, and the NengoLoihi package for compiling the model to run on Intel's Loihi neuromorphic chip \cite{davies2018loihi}. 

We illustrate two variations on this workflow by describing two example neurorobotics applications in detail.
The first example implements an end-to-end perception and action system in Nengo for tracking a target with a rover simulated in Mujoco~\cite{todorov2012mujoco}.
The rover has 4 mounted cameras whose input is fed into a DNN built using Keras.
The DNN estimates the distance to the target, and this estimate is sent to a control network which generates torques to apply to the steering wheel and drive wheels to move the rover to the target.
This full system is then compiled onto Loihi.
In the second example, we demonstrate how Nengo can be integrated with an existing system by augmenting a standard robotic arm force controller using a neural adaptive controller that learns online.  
We implement the adaptive component both on standard hardware and Loihi, where we take advantage of its on-chip learning.
We compare implementations of the adaptive control system as it drives a physical Jaco$^2$ robot arm from Kinova to perform a reaching task while adapting to the unmodelled force of holding a two pound weight.
We discuss the workflow bottlenecks and challenges that are encountered, addressed, and remaining.

\section{Background}

\subsection{Nengo and supporting development packages}

Nengo is a neural modeling development and simulation platform.
Users specify the architecture of models using a Python-based API, referred to as the `front-end', and then compile their model for simulation on hardware using a `back-end'.
The API is designed such that the same model can be run on different hardware with few to no changes in the front-end script.
Supported hardware includes CPUs, GPUs, FPGAs, and specialized neuromorphic hardware (such as Intel's Loihi chip).

Nengo users are able to quickly design and simulate neural networks, and use the NengoGUI package to visualize and interface with them during run-time.
The NengoDL package extends Nengo's API to interface with and integrate deep and machine learning networks built in Keras or TensorFlow, as well as take advantage of TensorFlow's resource distribution manager for efficient simulation across multiple processors.
The NengoInterfaces package provides easy interface access with the Mujoco simulator, abstracting out the setup and overhead involved in connecting, running, communicating, and restarting Mujoco simulations.
The NengoLoihi package allows us to compile our models to run on the Loihi, and also handles communication to and from the chip.
Additionally, the NengoLoihi package provides a Loihi emulator that allows users to run their models while simulating Loihi dynamics and computations on their computers, which aids efficient development.
Nengo has more supporting packages, but in the interest of space we limit our review to the above packages relevant to the work presented in this paper.

\subsection{The Loihi chip}
\label{sec:loihi}

We use Intel's Loihi chip~\cite{davies2018loihi} for demonstration in the examples below.
The Loihi chip is a many-core mesh with 128 neuromorphic cores, 3 embedded x86 processing cores, and off-chip communication interfaces.
An asynchronous network-on-chip communicates packetized messages between cores, allowing write, read request and response, spike messages, and barrier messages for time synchronization to be sent between cores.
Each neuromorphic core has 1,024 neural `compartments', where a compartment can be allocated to simulate a neuron or dendrite.

The NengoLoihi package allows users to compile their front-end script to the Loihi chip, handling all of the low-level mapping and communication.
Some front-end scripts will require modification specifying low-level details, such as how to allocate neural populations across Loihi cores, but largely the details of this low-level mapping are abstracted out and handled automatically.

\subsection{Other neurorobotic workflows and toolkits}

Most commonly, building neurorobotic applications involves hand-crafting and tuning SNNs for the task of interest.
In \cite{gutierrez2020neuropod} the authors build an SNN inspired by biology to implement a central-pattern generator that runs on the SpiNNaker neuromorphic board~\cite{furber2014spinnaker} and drives a hexapod robot to walk, trot, or run.
In \cite{kreiser2019self}, the authors hand craft an SNN to run on the Loihi to steer a small rover.
In \cite{stagsted2020towards}, a PID controller is implemented in an SNN running on Loihi to steer an unmanned aerial vehicle. 
The authors accomplish this using one-hot encoding, such that only one neuron in a population is able to spike at a time, and each neuron represents a different possible variable value, to build up networks implementing addition and subtraction, at which point a PID controller can be built.
Implementation of non-linear functions is listed as future work.
We note that these models can be built using the Nengo API, and the NEF API makes non-linear function implementation straight-forward.

The authors of \cite{taunyazov2020event} implement an SNN visual tactile system that runs on Loihi and performs container classification and rotational slip detection.
The network is a deep net trained with the SLAYER~\cite{bam2018slayer} method, which uses stochastic spiking neurons to overcome the undefined derivative in spiking neurons that prevents backpropagation from working.
In \cite{hwu2017self}, the authors train an Energy-Efficient Deep Neural Network (EEDN), a deep network designed specifically to run on IBM's TrueNorth neuromorphic chip~\cite{sawada2016truenorth}, on trail photos to train up a network that attaches to a real-world rover and guides it along a path.
The authors mention that the trained weights work well in a standard convolutional neural network or in the EEDN, which can transfer its weights directly to the TrueNorth.
As we detail in our examples below, the NengoDL package allows users to take advantage of similar deep learning methods for training SNNs to run on neuromorphic hardware.

Another way to program neuromorphic hardware is using the Python Neural Networks (PyNN) interface~\cite{davison2009pynn}. 
PyNN is a front-end API that shares Nengo's goal of creating a high-level front-end API that specifies neural network architecture without being tied to the low-level implementations specifics.
PyNN was developed to standardize scripting neural networks across several different low-level neural simulators, including Brian~\cite{goodman2008brian}, NEURON~\cite{hines1997neuron}, and Nest~\cite{gewaltig2007nest}.
Since its development, others have extended PyNN to include backends for other simulators and neuromorphic hardware. 
While Nengo allows the same low-level specificity of PyNN, Nengo also allows many of these details to be easily abstracted, and has more focus on high-level objects that speed the development of large or complex neural systems.
For example, to implement a communication channel between two populations of neurons takes approximately 12 lines in Nengo, and over 80 lines in PyNN~\cite{bekolay2011learning}.
Importantly, Nengo also supports methods for generating connection weight matrices, including the NEF as well as Keras/TensorFlow techniques, which PyNN does not.

The Neurorobotics Platform (NRP; ~\cite{falotico2017connecting}) is a web-based simulation environment for running SNNs hooked up to virtual robots.
Developed as part of the Human Brain Project~\cite{markram2012human}, the goal of the NRP is to streamline the process of running experiments with SNNs and robots.
The NRP lets users quickly select a virtual environment, robot, and SNNs to run an experiment. 
The NRP has a broad scope, offering tools such as the web-based robot designer, experimental workflow editor, and Gazebo simulation environment editor.
In contrast, Nengo focuses primarily on the development, integration, and simulation of neural networks (both spiking and non-spiking), support of different neural network programming paradigms like the NEF and Keras/TensorFlow, compiling to different hardware backends, and systems interfacing through Python.
There is potential for collaboration between Nengo and the NRP, expanding the neural network development and simulator interfacing of the NRP, and the experiment design and web interface of Nengo.

\section{Neurorobotic rover system}
\begin{figure}
    \centering
    \includegraphics[scale=.67]{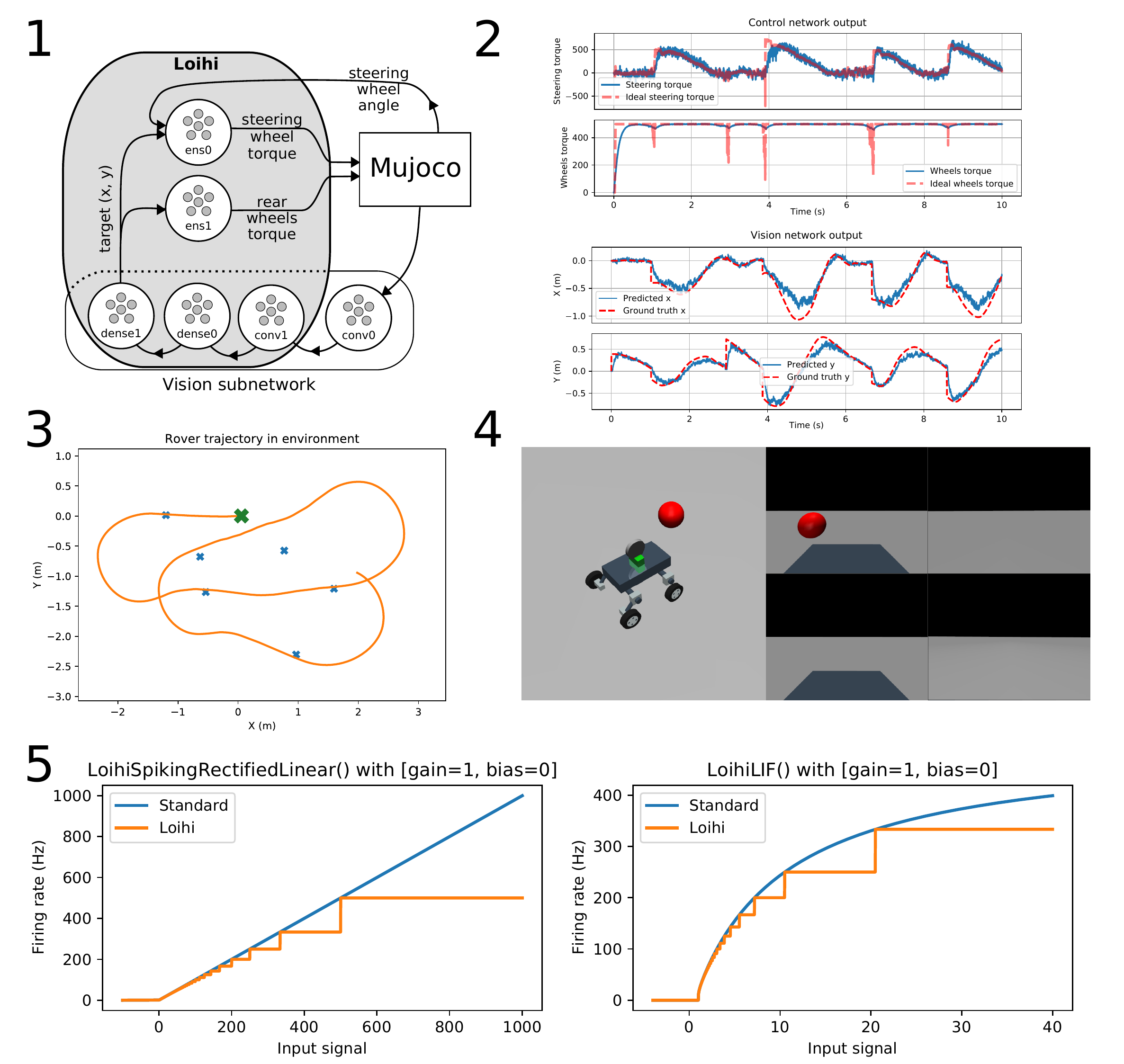}
    \caption{A neurorobotic rover. 1) A diagram of the neural network implementing the rover's perception and action system. The components inside the gray rounded rectangle are run on the Loihi hardware. Arrows indicate the flow of information. The `conv' prefix denotes convolutional layers, the `dense' prefix denotes fully connected layers, and the `ens' prefix denotes NEF layers. 2) Top: The control network's approximation of the steering (top) and acceleration (bottom) signals are plotted in blue. The ideal function output is plotted as a dashed red line. Bottom: The vision network's estimate of the target x (top) and y (bottom) location relative to the rover is plotted in blue. Ground truth is plotted as a dashed red line. 3) The trajectory followed by the rover is plotted in orange as it approaches different targets, as seen from above. The rover starts in the center at the green X and drives to the different targets, plotted as blue Xs. 4) Left: An image of the rover and the target in the Mujoco environment. Right: The images from the four mounted cameras attached to the rover, which generate the input to the vision network. 5) The firing rate curve of standard and Loihi neurons, to show the effects of discretization on the on-chip activity profiles neurons. Left) Spiking rectified linear neurons; right) Leaky integrate-and-fire neurons.}
    \label{fig:rover}
\end{figure}
In this example we develop an end-to-end perception and action system for tracking a target with a rover in Mujoco \cite{todorov2012mujoco}. 
The simulated rover we use is a 4 wheeled vehicle, built using Mujoco's XML modeling language, with Ackerman steering and rear differential drive in a boundless environment with no obstacles.
The rover has 4 RGB cameras mounted on its back, each with a 90 degrees field-of-vision, that provide a full 360 degrees view of the environment, and a sensor on the front wheels that provide steering angle information.
The rover accepts two torque input signals, one to control the acceleration of the rear wheels, and one to turn the front wheels right or left. 
The target is a red sphere that floats in the air and warps to a new location (generated from a random distribution within 3 metres of the origin) when the centre-of-mass of the rover is within 50 centimetres of the center of the target.
Figure~\ref{fig:rover}-4 shows the rover in the world on the left, and the view from each of the 4 cameras on the right.

To begin developing our perception and action systems, we first build out the controller without using neural networks. 
We use the exact ground-truth information provided by Mujoco to identify the target location relative to the rover, and calculate the torques for acceleration and steering in Python.
Next, we address sensory perception by building a DNN that can identify target $(x, y)$ location relative to the rover based on input from the four mounted cameras.
This requires building a dataset using the simulator to train the DNN.
We initially train the system using standard artificial rate neurons, to confirm that the desired level of performance can be achieved with our network architecture.
We then us NengoDL to convert the network to spiking neurons and tune the parameters to optimize performance.
Next, we replace each function in the control system with spiking analogues, testing each in isolation before finally integrating the entire spiking network.
Once performance is achieved in a fully spiking neural network, we move the network simulation from Nengo into the NengoLoihi emulator, and tune the parameters again to optimize under the constraints of the Loihi.  
Finally we compile the network to the Loihi hardware, again using NengoLoihi.  
In the next sections we describe each of these steps in more detail.

\subsection{Interfacing with Mujoco}

Interfacing to Mujoco is done through NengoInterfaces, which uses the \texttt{mujoco-py} \cite{mujocopy} library for Python bindings to the Mujoco C API.
The interface accepts force signals from the neural network, applies them inside Mujoco and moves the simulation forward one time step, and then returns feedback from the rover.
Environment information can be accessed directly from the NengoInterfaces API, and less common functions are available through the \texttt{mujoco-py} simulation and environment model parameters.

\subsection{Processing visual input using a Keras DNN converted to a Nengo SNN}

The network used for tracking the target location consists of 2 convolutional layers and 3 dense layers, and is shown in the bottom block of Figure~\ref{fig:rover}-1.
The first convolutional layer uses 1x1 kernels with a stride of 1, and a filter size of 3; its purpose is to convert the image signal into spikes to be sent to the layers running on Loihi \footnote{It is also possible to send information to Loihi by setting the bias and current for neurons, but we have found this method to be slower for a dynamic input signal.}.
To generate the input image, we take a 32x32 pixels resolution snapshot from each camera, and concatenate them horizontally to create a 32x128 pixels input to the network.
This resolution was chosen as the smallest network size that could still identify targets at a distance of 3 metres.
To retrieve the signal from the last layer running on-chip we use neural probes, which monitor spiking activity. 

The dataset used for training the model was generated by recording both input from the mounted cameras and the relative distance to the target. 
The data was collected while our non-spiking  control system drove the rover to the targets, recording every 10th frame. 
The final dataset used consists of roughly 40,000 images and target (x, y) locations (the height of the target is constant and is not relevant to control so we ignore it).

Training the network with non-spiking ReLU activation functions using standard DNN tools (i.e., Keras) was the first step.
This allows us to validate the network architecture.
For all of our training we use the RMSprop optimizer from TensorFlow and the mean squared error loss function on network output to learn to output the target (x, y) locations associated with an image.
When converting the network into spiking neurons take into account both the desired firing rates and the activation function of neurons running on the Loihi.
We have found that if the average firing rates are less than 50Hz, spiking neurons are not driven strongly enough to generate any activity.
We target the 175Hz range for firing rates, because it is large enough to ensure spiking, but still inside the range where the Loihi neurons approximate standard neurons well (discussed below).
To achieve this, we initialize the weights of our network by setting the \texttt{scale\_firing\_rates} parameter of the NengoDL built-in Keras converter to \texttt{400}. 
This parameter encourages the optimizer to converge to firing rates that are higher or lower, based on the scaling.
An alternative, and more fine-grained and reliable method, is to add a firing rate regularization term to the cost function that penalizes neurons firing outside of the desired range.

The second factor we need to account for is the activation function of neurons on the target neuromorphic hardware.
Neurons on the Loihi have a unique activation profile because of discretization that occurs on-chip, as shown in Figure~\ref{fig:adaptive_control_results}-5.
We use NengoLoihi's model of the Loihi rectified linear neuron during training to ensure that the network is trained on the same kind of activation functions used during inference.

Finally, we set network synapses throughout the network as required to smooth out the signal and filter noise.
In Nengo, you can set synapses to 0 to implement no filtering, or to None to collapse the computations of two connected layers into a single layer.
We set each of the synapses on the connections between layers to None.
This speeds up the propagation of spikes through the network, but also has the potential to decrease performance due increased noise in the signal.
We found empirically, however, that applying a 0.05s time constant low-pass filter on the vision network output smoothed the target location estimate and improved the control signal generated downstream. 

\subsection{Generating robust, explainable control signals using the NEF}
\label{sec:NEF}

We described the NEF as a `white-box' approach to building neural networks because of its mechanistic approach.
Briefly, the NEF uses populations of neurons to represent vectors, feed-forward connections between populations to implement functions on those vectors, and recurrent connections to implement differential equations.
Rather than specifying a task-level cost function, as in standard machine learning methods, the user must first design a circuit that solves the problem, including specifying a state space representation, set of computations, and flow of information.
The user then uses Nengo's NEF API to implement this circuit in neurons.
These added top-down constraints give clear network structure that allows users to identify points of error and apply specific changes to debug and improve network performance.
This is in contrast to `black-box' deep learning methods, which use training algorithms to find a network configuration that solves the problem, without knowledge of the function implemented.
In situations such as complex visual analysis where the algorithmic solution is unknown this is very desirable, but in cases where proven solutions are available and we would like performance guarantees the methods of the NEF are preferable.

The motor system for the rover is implemented using the NEF.
The first step in this implementation is deriving the functions for vehicle acceleration and steering wheel control. 
We calculate acceleration as distance to target multiplied by a gain term, clipped to a maximum magnitude empirically chosen to prevent the rover from flipping when travelling at top speed and turning sharply.
Formally,
\begin{equation}
    u_{acceleration} = k_a \textrm{min}(\|(x^*, y^*)\|, 1)
\end{equation}
where $u$ is the control signal sent to the rover, $k_a$ is the acceleration gain term, $x^*$ and $y^*$ are the target location relative to the rover, and $\|\cdot\|$ denotes the 2-norm.

The torque applied to the steering wheel is calculated with a simple proportional controller using the difference between the current and desired angle of the wheels multiplied by a gain term.
The desired angle is calculated as the angle to the target using $\textrm{arctan2}(-x^*, y^*)$, where order of the arguments and the negative sign in front of the $x$ term account for the orientation of the rover relative to the environment.
Formally,
\begin{equation}
    u_{steer} = k_p (\textrm{arctan2}(-x^*, y^*) - q)
\end{equation}
where $k_p$ is the proportional gain term, and $q$ is the current angle of the steering wheel.

The neural circuit implementation of this controller is done in two parallel ensembles, with the connections weights on the outbound connections calculated to approximate the above equations, as shown in the top portion of Figure~\ref{fig:rover}-1.
In Nengo, we project the relevant variables into each population and specify the functions to be computed on their outbound connections using Python code.
Nengo then solves for connection weights using the principles of the NEF. 
In this particular case, we compute these functions using separate ensembles, rather than having a single ensemble with two separate outbound connections, because the functions they are calculating depend on different sets of variables.
The acceleration function only requires the estimated target (x, y) values for calculation, while the steering function requires the estimated target (x, y) and the current angle of the steering wheel.
While the variables required by the acceleration function are a subset of the variables used in the steering function, we can achieve greater precision by using an ensemble that only encodes the target (x, y).

In the hardware implementation, each ensemble consists of 4,096 Loihi leaky integrate-and-fire neurons, spread across 4 cores.
This specific number of neurons is chosen to satisfy hardware constraints on the number of inbound connections a population of neurons can receive.
The neurons in the ensemble are set up to have maximum firing rates between 175 and 220Hz, chosen because in general higher firing rates provide for more accurate function approximation.  

\subsection{Integration and compiling to hardware}

Putting the vision and control networks together is a simple matter of connecting the output of one to the input of the other in Nengo.
As both networks were built using Loihi-type neurons, they are also prepared to be mapped to neuromorphic hardware.
During the initial building and debugging process running on a CPU backend, which is the Nengo default, greatly expedited development.
The NengoLoihi backend can then be used to compile the network to run on the Loihi (we could also use NengoOCL or NengoDL to compile to GPU and run directly in Nengo).
We used a workstation with and Intel Core i7-6700K CPU @ 4.00 GHz x 8 with 32 GB RAM and GeForce GTX 1070/PCIe/SSE2 running Ubuntu 18.04, and the Intel Nahuku board with 32 Loihi chips \cite{davies2018loihi} running NxSDK 0.9.
In this example, we are only using one of the Loihi chips on the board.
When the system is running, Nengo provides the interface between the simulator and the hardware, but all computations are run on the Loihi chip.

\subsection{Performance}

Figure~\ref{fig:rover}-3 shows the (x, y) trajectory of the rover moving throughout the environment to 6 different targets in the environment, starting from the green `x'.
Figure~\ref{fig:rover}-2 shows the perception and action signals from the network, with the target (x, y) estimated in the top figures in blue and the ground truth shown in red.
The lower figure shows the steering and acceleration control signals generated by the network in blue with the ideal values in orange.
As can be seen, the rover drives accurately (to within 50cm) over the course of the trial.
We have not performed extensive testing of the accuracy, and do not provide quantitative results as our purpose here is to focus on the methods used to develop the system.
There is clearly significant room for improvement and extension to this work.
Nevertheless, this simple example demonstrates the implementation of an end-to-end perception and action spiking neural network running on neuromorphic hardware.
All code is available online at \url{https://github.com/abr/neurorobotics-2020}.
We have provided full code to serve as a starting point for those interested in exploring neurorobotic solutions that can leverage embedded neuromorphic hardware for next generation systems. 

\section{Neurorobotic adaptive arm control}
\begin{figure}
    \centering
    \includegraphics[scale=.205]{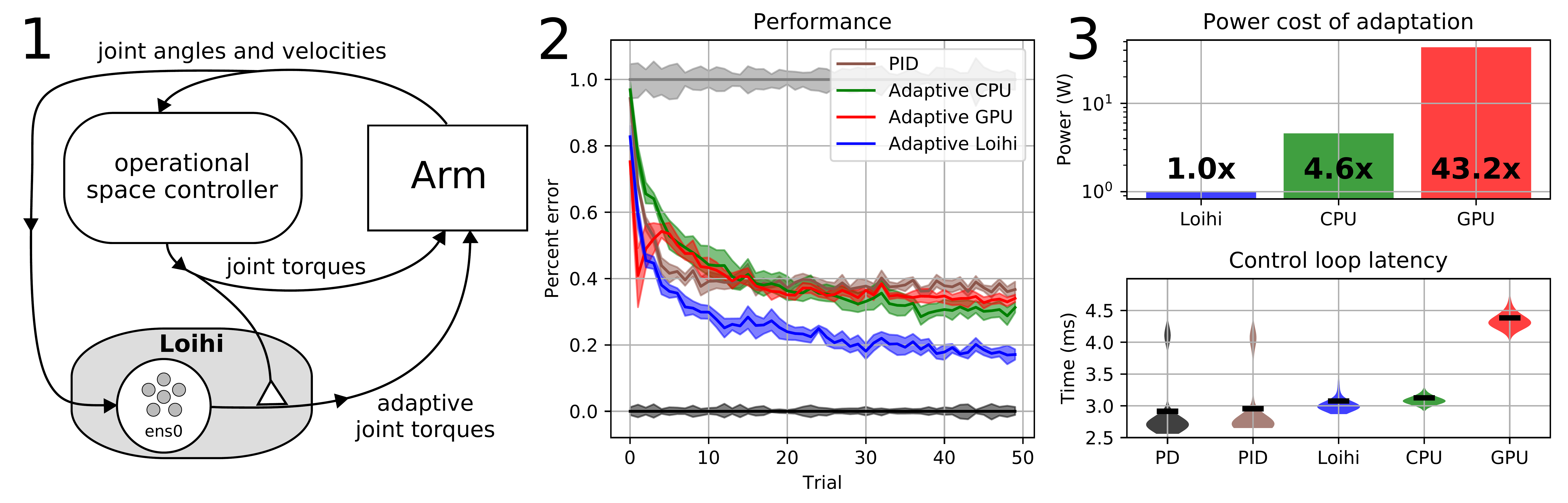}
    \caption{
    A neurorobotic adaptive arm controller. 
    1) The system diagram of the neurorobotic adaptive control system. 
    The hollow triangle denotes the connection providing the training signal to the learning rule. 
    Note that Nengo is only used to run the neural part of the system. 
    The `ens' prefix denotes NEF layers. 
    2) Performance error while reaching to a target while holding an unexpected two pound mass. 
    Black: PD controller with no extra mass, used as a 0\% reference error. 
    Grey: PD controller reaching while holding the two pound mass, used as 100\% reference error. 
    Brown: PID controller reduces error to 39.1\%. 
    Green: Adaptive controller CPU implementation with 1,000 neurons reduces error to 37.6\%. 
    Red: Adaptive controller GPU implementation with 1,000 neurons reduces error to 36.2\%. 
    The initially high error is due to the increased latency of the GPU implementation. 
    Blue: Adaptive controller Loihi implementation with 1,000 neurons reduces error to 26.7\%. 
    Results are averaged over five sets of 50 trials with 95\% confidence intervals shown. 
    The adaptive controller demonstrates a 2.45 times improvement in accuracy over PID, and 1.49 and 1.57 times improvement over the CPU and GPU adaptive controller implementations, respectively. 
    3) Top: A power comparison between adaptive controller implementations. 
    Running adaptation on the CPU and GPU requires 4.6x and 43.2x more power than Loihi. 
    Bottom: Latency measurements of the controllers. 
    PD: 2.91ms, PID: 2.95ms, Adaptive Loihi: 3.08ms, Adaptive CPU: 3.13ms, Adaptive GPU: 4.38ms
    }
    \label{fig:adaptive_control_results}
\end{figure}
In this second example we augment an existing force controller with an adaptive neural network implemented on Loihi, using on-chip learning to control a Kinova Jaco$^2$ physical robot in a reaching task while it holds an unexpected weight.
The existing control system generates joint torques using a standard proportional derivative (PD) operational space controller (OSC; \cite{slotine1988robust}), designed to move the hand along a target path.  
The adaptive controller adds an adaptive signal trained online to account for any unexpected forces affecting movement, which is tuned online.
We compare performance of the adaptive control system to a non-adaptive PD OSC, and an industry standard proportional integrated-error derivative (PID) OSC.
We also compare the neuromorphic implementation with CPU and GPU implementations of the adaptive control system. 
We compare all systems in terms of accuracy, power use, and control loop update latency. 

Operational space control relies on an accurate model of the arm dynamics to generate torques that will move the arm as desired.
If the arm picks up an object, is subjected to external forces or perturbances, or wears down over time, the dynamics have changed and OSC performance will degrade unless the changes can be accurately modeled.
In general, these perturbations can not be predicted in advance, so updating the controller on-the-fly is desirable.
This is the purpose of the adaptive controller we use here, implemented as a neural network.
Intuitively, the adaptive controller acts as a context sensitive integrated error term.
Where standard integrated error terms (such as the I in PID control) apply the same learned error regardless of the current joint angles or velocities, the adaptive controller learns to account for errors specific to different arm states.
The difference becomes significant in the control of highly nonlinear systems, where the error that needs to be compensated for changes significantly with system state.

\subsection{Interfacing with the Jaco2 robotic arm}

In this example Nengo is called as a sub-function of the PD OSC.
The PD OSC itself is implemented in Python and runs on a workstation that interfaces with a physical Kinova Jaco$^2$ 6 degrees-of-freedom (DOF) arm. 
The interface implemented is through the ABR Jaco$^2$ repository, and includes no neural network infrastructure.  
To integrate neural computation with this standard Python code, at each time step in the control model, Nengo is called to run the neural network for a single step.
The control code sends feedback from the arm to Nengo and receives back an adaptive control signal to add into the outbound set of joint torques sent to the Jaco$^2$. 
Nengo takes care of running the neural network on a CPU, GPU, or the Loihi neuromorphic hardware.

\subsection{Processing sensory feedback using the NEF}

In this application we are augmenting an existing control system with an adaptive control signal generated by a neural ensemble. 
The ensemble requires sensory feedback related to joint positions and velocities as input in order to compute the necessary correction to the control signal.
Because the sensory feedback is a relatively low-dimensional signal (e.g., compared to image input), it does not need to be processed by a deep neural network before we can use it to generate a corrective control signal.
The adaptive controller requires that the effects of the unexpected force are predictable given the input provided to the ensemble to be able to learn to compensate for unexpected forces affecting the arm.
If, for example, the force affecting the arm was a function of joint angle, and the ensemble only had inputs related to joint velocity, then the network would not be able to adapt to the force.

Given that we have appropriate inputs, we further need to make sure that the neurons are sufficiently sensitive to different states of the arm relevant for compensating for the unexpected force.
In the NEF, the neural tuning properties are determined by a combination of the neuron encoders (or `preferred' direction vectors), gains, and biases.
This tuning determines which parts of the input state space are represented by the neural ensemble.  
In this section, we present considerations that determine how to appropriately pick these tuning curves for adaptive control in a highly non-linear state space.

In particular, we need to ensure that neurons are not active over a large part of state space. 
If this is the case, the compensatory signal they learn in one part of state space may incorrectly generalize to other parts.
In contrast, if neurons are active over a small part of state space, then the compensatory signal they learn in one area will not affect what learning occurs in other parts of state space.  
Unsurprisingly, there is a trade-off between the specificity of neural responses and their generalization abilities.
In arm control, because the dynamics are highly nonlinear, we generally want to ensure that neurons in our ensemble are sensitive to localized parts of state space. 
We also do not want to waste neural resources.  
Consequently, neurons should only be sensitive to parts of state space that are actually explored by the arm.  
In other words, we do not want our population to include neurons that never become active.  

To handle both of these issues, we need to carefully choose neural tuning curves, and hence NEF encoders. 
To optimize neurons for the relevant parts of the state space, we begin by subtracting the mean and then normalizing the input signals for each dimension given their joint limits.
Next, we project into the D+1 unit hypersphere, where D is the number of dimensions represented by the ensemble.
By doing this it becomes possible to carefully control the range of values that cause neurons in the ensemble to respond.
Because we have 6 joints and two input signals from each (i.e., position and velocity), we project the normalized signals onto the 13-dimensional (i.e., 12+1) unit hypersphere and use that as input to the ensemble.

\begin{figure}
    \centering
    \includegraphics[scale=.325]{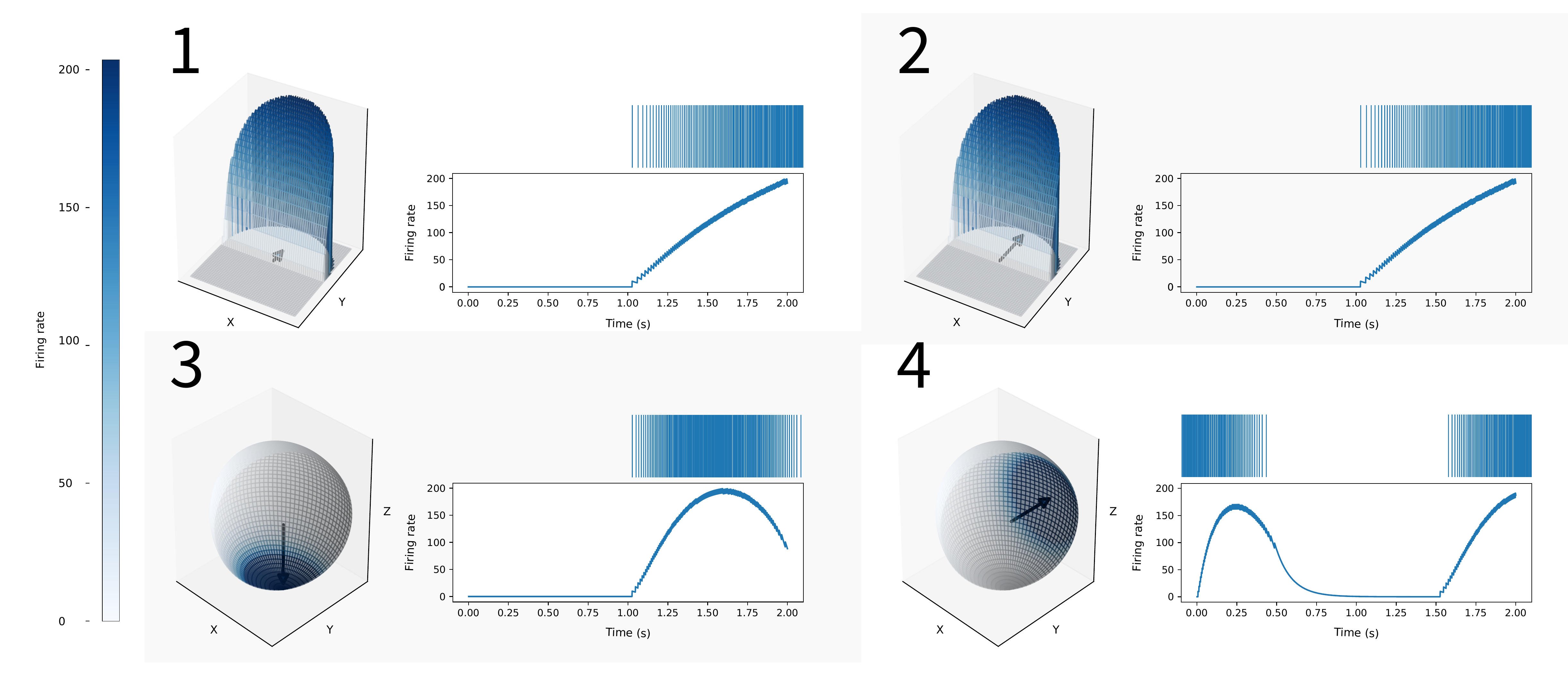}
    \caption{
    An example of how increasing encoder dimensionality can create neurons that are more selectively responsive. 
    On the right side of all figures the spike raster (top) and filtered output (bottom) is plotted, showing the firing rates of these neurons for the same 2D input signal that moves from [0, -1] to [0, 1], projected into 3D for 3 and 4.
    In the top row we show two neurons with 2D encoding vectors (0, 0.5) (shown in 1 on the left) and (0, 1.0) (shown in 2 on the left). 
    The arrows in the figures represent the neuron's encoding vectors.
    The z-axis and color reflect the firing rates of these neurons given different (x, y) input. 
    As can be seen, the activity of these neurons with co-linear encoding vectors is indistinguishable. 
    In the bottom row, we project these same 2D encoders, (0, 0.5) (shown in 3 on the left) and (0, 1.0) (shown in 4 on the left), into 3D. 
    The arrows in the figures represent the neuron's encoding vectors projected into 3D.
    The firing rates of the neurons given different (x, y, z) input is represented by color, showing that the neurons are responsive to different parts of 3D state space. 
    }
    \label{fig:encoders}
\end{figure}

Figure~\ref{fig:encoders} illustrates how this allows us to control neural responses by considering the simpler case of projecting 2D into 3D.
Assuming we have inputs in the -1 to 1 range along each dimension, the inputs are going to lie somewhere in the unit square.
In the NEF, by default, each of our neurons will initially have an encoder that is a vector pointing from the origin to somewhere inside unit circle.
Neurons with encoding vectors that point in the same direction will have similar firing rates responses to the same input, as shown in the top half of Figure~\ref{fig:encoders}.
By projecting our encoding vectors and input signal into the 2+1 dimensional unit hypersphere, our neurons are still sensitive to all parts of the original 2D input signal, but co-linear encoding vectors can also generate distinct activity, as shown in the bottom half of Figure~\ref{fig:encoders}.
For example, we can have neurons sensitive to an input signal of (0, 0.5) but not (0, 1), which was not possible with our 2D preferred direction vectors.
Essentially, this method allows us to have neurons sensitive to more specific parts of state space.

\subsection{Online learning for adaptive control}

The neural adaptive controller presented here is a neuromorphic implementation of the control system presented in \cite{dewolf2016spiking}, where it is proven that this adaptive controller performs as well or better than a PID controller. 
The neural adaptive controller uses the Prescribed Error Sensitivity (PES; \cite{macneil2011fine}) learning rule, which is a local, spiking or non-spiking, error-driven Hebbian rule.
The Loihi chip supports several different kinds of online learning, providing the ability to use microcode to define different kinds of rules.  
NengoLoihi implements the PES learning rule using this feature of the chip, which allows weight updates to be calculated on-chip.  
For the other hardware, core Nengo includes a definition of the PES rule.

The adaptive control signal is calculated via
\begin{equation}
    \textbf{u}_\textrm{adapt} = \textbf{a} \hspace{.5mm} \textbf{d},
\end{equation}
where $\textbf{a}$ is the vector of neural activities (i.e., filtered neural spike trains) and $\textbf{d}$ denotes a vector of `decoders' which are the output weights from the ensemble.
The resulting $\textbf{u}_{adapt}$ is a vector of the same dimensionality as the OSC control signal.
We initialize $\textbf{d}$ to a vector of zeros, and use the learning rule 
\begin{equation}
    \Delta \textbf{d} = -\kappa \; \textbf{a} \otimes \textbf{u},
\end{equation}
to update the decoder weights, where $\kappa$ is a learning rate, $\textbf{u}$ is the OSC's outbound control signal (acting as the error in the PES rule), and $\otimes$ denotes the outer product.
This training signal, $\textbf{u}$, was chosen based on Lyapunov stability analysis.
Details, derivation, and proof of stability of this adaptive neural controller are provided in \cite{dewolf2016spiking}.

In the system diagram shown in Figure~\ref{fig:adaptive_control_results}-1,  the hollow triangle denotes the connection providing the training signal, $\textbf{u}$, for the learning rule.

\subsection{Compiling to neuromorphic hardware}
\label{compiling}

The NengoLoihi backend is used to instantiate the neural ensemble on the Loihi with the parameters discussed above, and implement the PES learning rule on-chip.  Core Nengo is used for the CPU implementation and the NengoOCL backend package is used for the GPU implementation.
In this example we used a workstation with and Intel Core i7-6700K CPU @ 4.00 GHz x 8 with 32 GB RAM and GeForce GTX 1070/PCIe/SSE2 running Ubuntu 18.04 and the Intel Kapoho Bay board with 8 Loihi chips running NxSDK 0.9
In this example, we are only using one of the Loihi chips on the board.

\subsection{Performance}

Figure~\ref{fig:adaptive_control_results}-2 shows the arm performing a reaching task while holding an unexpected two pound mass.  
In this task the arm repeatedly starts from the same position and reaches to the same target 50 times, with continuous learning between reaches.
We perform the 50 reaches with each controller 5 times, using different randomly generated neuron ensemble parameters (such as bias and maximum firing rates), and calculate the mean error and 95\% confidence intervals.
The adaptive controller is run while simulating the neurons on the Loihi, and the CPU and GPU of our workstation (specifications are in Section \ref{compiling}), and compared against a standard PID controller running on the same workstation. 
We normalize our performance results using the performance of a PD operational space controller.
We consider that controller reaching under normal conditions with nothing in the hand as 0\% error, and the results of that controller reaching while holding the unaccounted-for two pound weight as 100\% error.  
As can be seen, the Loihi system outperforms all other controllers after 50 trials of training. 
Unsurprisingly, the CPU and GPU perform similarly, and better than the PID controller.

Figure~\ref{fig:adaptive_control_results}-3 shows the power and latency measurements of the controllers during the task, where the neuromorphic implementation consumes the least energy of the adaptive controllers, with a minimal increase in latency compared to the non-adaptive controllers.  
Specifically, the CPU uses 4.6x more power, and the GPU 43.2x more. 
As well, the CPU is a similar latency (i.e., 2\% slower), while the GPU is 42\% slower than the Loihi. 

To measure the power use of the CPU, we used the software package \texttt{s-tui}, available online at \url{https://github.com/amanusk/s-tui}.
To measure the power use of the GPU, we used the \texttt{nvidia-smi} software.
To measure the Loihi power use, we used the Linear Tech DC1613A dongle and LTpowerPlay software, which provides current and voltage measurements for the chip's two power supplies, from which we calculated the total power use.

\section{Discussion}

We have demonstrated how to take advantage of neuromorphic technology to fully implement or augment existing robotic control systems.
In particular, we showed how a set of tools in the Nengo ecosystem allows efficient execution of four central tasks for building neurorobotic systems.

\begin{enumerate}
    \item %Developing infrastructure to send and receive signals from the environment.
    The NengoInterfaces library provides an easy API for interfacing with the Mujoco simulation, used in the first example, both for sending in control signals and receiving feedback.
    While the second example is not directly providing simple API access, it illustrates the flexibility of Nengo to be incorporated into already developed Python programs and interfaces.
    \item %Processing task specific sensory signals.
    In the rover example, we show how NengoDL provides a natural way to integrate Keras and TensorFlow models.
    For non-neural perception algorithms, Python code can be directly executed from Nengo or the code can be run outside Nengo and sent into a Nengo model, as is done in the adaptive arm example.
    \item %Generating robust control signals with explainable neural networks.
    We illustrated how a circuit design that solves the problem of interest can be implemented in a neural network using Nengo, providing white-box neural network systems.
    In the first example this was shown with the rover control network that steered and drove the rover to the target, and the second example showed the use of a vector-space training signal with stability guarantees to implement non-linear adaptive control.
    \item %Compiling neural networks to run on multiple targeted hardware platforms.
    The Nengo development toolkit allows users to compile their model to run on multiple different hardware platforms, including CPU~\cite{bekolay2014nengo}, GPU~\cite{rasmussen2019nengodl}, FPGA~\cite{morcos2019nengofpga}, Intel's Loihi~\cite{nengoloihi}, and SpiNNaker~\cite{mundy2015efficient}.
\end{enumerate}

We have made all of the code used in these examples publicly available, to provide practical, reproducible examples for the community.
We believe this set of tools and examples helps address the core challenge of making neurorobotic systems easier to build, and a wide variety of architectures easier to explore.

In the first example, while we did not benchmark or quantitatively characterize the result, it provides a demonstration of how our chosen tools allow the development of complete perception-action systems for neurorobotics.
In particular, it demonstrated how to couple white box (i.e., NEF) and black box (i.e., DNN) techniques and implement them on a single underlying spiking neuromorphic hardware platform (i.e,. Loihi). 

Nengo is unique in its ability to make such integration easily accessible.
Nengo has several advantages in this context: a) it is not vendor specific, and supports hardware from several sources; b) it allows a high-level model specification for easy portability, while also allowing hardware specific details to be incorporated (e.g., via its configuration system); and c) Nengo removes the need to have detailed knowledge about SNNs, neuromorphic hardware, simulator interfaces, embedded programming, and so on, while allowing those with such knowledge to leverage it (e.g., by easily defining new neuron models, using the configuration system, building new hardware specific backends, and so on). 

Our second example provided a more quantitative characterization of the advantages of neurorobotics.
From a tools perspective, it demonstrated how Nengo can be integrated into existing systems and run the same neural model across multiple kinds of hardware.
But, more importantly, this example demonstrates the kinds of advantages we expect from neuromorphics: an increase in speed and accuracy, and several-fold decrease in power compared to traditional hardware.
As is well established, low latency and energy efficiency are critical for many mobile robotics applications.  

Possible extensions to the examples provided here are many and varied.
Perhaps one of the more obvious ones is to implement the entire adaptive controller on neuromorphic hardware.
To the best of our knowledge this has only been done with a 3-link planar arm~\cite{dewolf2016spiking}. 
Since the adaptive controller has performance guarantees, and the white box methods of the NEF allow us to implement it on neuromorophic hardware while preserving those guarantees, a full implementation would be a rare example of an adaptive, fully neurorobotic controller with clear performance guarantees.

While we believe the Nengo ecosystem is useful for the development of neurorobotic systems, there remain a variety of challenges and directions for future development that stand to improve it.
For instance, Nengo backends that target non-spiking AI acceleration hardware, such as Google's Coral chip, would expand the community able to use the methods we have discussed because spiking neuromorphic hardware, such as Intel's Loihi chip, is not commercially or otherwise widely available.
Extending the interfaces offered by the NengoInterfaces package to improve accessibility, as well as offering the same automatic conversion from DNNs to SNNs for PyTorch users also remains important future work. 
Perhaps most importantly, continuing to increase the number of available tutorials, ready-to-use models, and online examples is critical to reducing the startup overhead for new users and better supporting the neurorobotics, neural networks, and edge AI communities.

In conclusion, the Nengo ecosystem makes it possible for users to quickly develop applications for neuromorphic hardware, while taking advantage of already developed neural or non-neural machine learning solutions.
We have shown two examples that demonstrate how the ecosystem can be used to address four core stages of the development workflow. 
We encourage interested researchers to use the code and tools that we have made available, and look forward to exploring the vast space of robust, embedded neurorobotics systems with the emerging research community.

\section*{Conflict of Interest Statement}
%All financial, commercial or other relationships that might be perceived by the academic community as representing a potential conflict of interest must be disclosed. If no such relationship exists, authors will be asked to confirm the following statement: 

This work was funded by Applied Brain Research, Inc and Intel. Nengo and related packages belong to Applied Brain Research, but are freely available for non-commercial use.

\section*{Author Contributions}

TD directed the project, TD, PJ, and CE implemented the examples, PJ and CE collected data for the arm example, TD wrote the manuscript, CE provided research guidance and edited the manuscript.

\section*{Funding}

The NengoLoihi interface development was funded by Intel. All other funding was provided by Applied Brain Research, Inc.

\section*{Acknowledgments}

The authors would like to thank Daniel Rasmussen, Eric Hunsberger, and Xuan Choo for their help in the development of this project.
We would also like to thank Intel for access to the Nahuku board and Kapoho Bay used in the examples.

\section*{Data Availability Statement}
The code and datasets generated and analyzed for the neurorobotic adaptive arm control example can be found at \url{https://www.github.com/abr/neurorobotics2020}.

\bibliographystyle{unsrt}
\bibliography{nengo_neurorobotics}

\begin{thebibliography}{10}

\bibitem{mead1990neuromorphic}
Carver Mead.
\newblock Neuromorphic electronic systems.
\newblock {\em Proceedings of the IEEE}, 78(10):1629--1636, 1990.

\bibitem{eliasmith2004neural}
Chris Eliasmith and Charles~H Anderson.
\newblock {\em Neural engineering: Computation, representation, and dynamics in
  neurobiological systems}.
\newblock MIT press, 2004.

\bibitem{bekolay2014nengo}
Trevor Bekolay, James Bergstra, Eric Hunsberger, Travis DeWolf, Terrence~C
  Stewart, Daniel Rasmussen, Xuan Choo, Aaron Voelker, and Chris Eliasmith.
\newblock Nengo: a python tool for building large-scale functional brain
  models.
\newblock {\em Frontiers in neuroinformatics}, 7:48, 2014.

\bibitem{davies2018loihi}
Mike Davies, Narayan Srinivasa, Tsung-Han Lin, Gautham Chinya, Yongqiang Cao,
  Sri~Harsha Choday, Georgios Dimou, Prasad Joshi, Nabil Imam, Shweta Jain,
  et~al.
\newblock Loihi: A neuromorphic manycore processor with on-chip learning.
\newblock {\em IEEE Micro}, 38(1):82--99, 2018.

\bibitem{todorov2012mujoco}
Emanuel Todorov, Tom Erez, and Yuval Tassa.
\newblock Mujoco: A physics engine for model-based control.
\newblock In {\em 2012 IEEE/RSJ International Conference on Intelligent Robots
  and Systems}, pages 5026--5033. IEEE, 2012.

\bibitem{gutierrez2020neuropod}
Daniel Gutierrez-Galan, Juan~P Dominguez-Morales, Fernando Perez-Pe{\~n}a,
  Angel Jimenez-Fernandez, and Alejandro Linares-Barranco.
\newblock Neuropod: a real-time neuromorphic spiking cpg applied to robotics.
\newblock {\em Neurocomputing}, 381:10--19, 2020.

\bibitem{furber2014spinnaker}
Steve~B Furber, Francesco Galluppi, Steve Temple, and Luis~A Plana.
\newblock The spinnaker project.
\newblock {\em Proceedings of the IEEE}, 102(5):652--665, 2014.

\bibitem{kreiser2019self}
Raphaela Kreiser, Gabriel Waibel, Yulia Sandamirskaya, et~al.
\newblock Self-calibration and learning on chip: towards neuromorphic robots.
\newblock In {\em IEEE/RSJ International Conference on Intelligent Robots and
  Systems (IROS 2019)}, 2019.

\bibitem{stagsted2020towards}
Rasmus~Karn{\o}e Stagsted, Antonio Vitale, Jonas Binz, Leon~Bonde Larsen, Yulia
  Sandarmirskaya, et~al.
\newblock Towards neuromorphic control: A spiking neural network based pid
  controller for uav.
\newblock In {\em Robotics: Science and Systems 2020}, 2020.

\bibitem{taunyazov2020event}
Tasbolat Taunyazov, Weicong Sng, Hian~Hian See, Brian Lim, Jethro Kuan,
  Abdul~Fatir Ansari, Benjamin~CK Tee, and Harold Soh.
\newblock Event-driven visual-tactile sensing and learning for robots.
\newblock {\em perception}, 4:5, 2020.

\bibitem{bam2018slayer}
Sumit Bam~Shrestha and Garrick Orchard.
\newblock Slayer: Spike layer error reassignment in time.
\newblock {\em arXiv}, pages arXiv--1810, 2018.

\bibitem{hwu2017self}
Tiffany Hwu, Jacob Isbell, Nicolas Oros, and Jeffrey Krichmar.
\newblock A self-driving robot using deep convolutional neural networks on
  neuromorphic hardware.
\newblock In {\em 2017 International Joint Conference on Neural Networks
  (IJCNN)}, pages 635--641. IEEE, 2017.

\bibitem{sawada2016truenorth}
Jun Sawada, Filipp Akopyan, Andrew~S Cassidy, Brian Taba, Michael~V Debole,
  Pallab Datta, Rodrigo Alvarez-Icaza, Arnon Amir, John~V Arthur, Alexander
  Andreopoulos, et~al.
\newblock Truenorth ecosystem for brain-inspired computing: scalable systems,
  software, and applications.
\newblock In {\em SC'16: Proceedings of the International Conference for High
  Performance Computing, Networking, Storage and Analysis}, pages 130--141.
  IEEE, 2016.

\bibitem{davison2009pynn}
Andrew~P Davison, Daniel Br{\"u}derle, Jochen~M Eppler, Jens Kremkow, Eilif
  Muller, Dejan Pecevski, Laurent Perrinet, and Pierre Yger.
\newblock Pynn: a common interface for neuronal network simulators.
\newblock {\em Frontiers in neuroinformatics}, 2:11, 2009.

\bibitem{goodman2008brian}
Dan~FM Goodman and Romain Brette.
\newblock Brian: a simulator for spiking neural networks in python.
\newblock {\em Frontiers in neuroinformatics}, 2:5, 2008.

\bibitem{hines1997neuron}
Michael~L Hines and Nicholas~T Carnevale.
\newblock The neuron simulation environment.
\newblock {\em Neural computation}, 9(6):1179--1209, 1997.

\bibitem{gewaltig2007nest}
Marc-Oliver Gewaltig and Markus Diesmann.
\newblock Nest (neural simulation tool).
\newblock {\em Scholarpedia}, 2(4):1430, 2007.

\bibitem{bekolay2011learning}
Trevor Bekolay.
\newblock Learning in large-scale spiking neural networks.
\newblock Master's thesis, University of Waterloo, 2011.

\bibitem{falotico2017connecting}
Egidio Falotico, Lorenzo Vannucci, Alessandro Ambrosano, Ugo Albanese, Stefan
  Ulbrich, Juan~Camilo Vasquez~Tieck, Georg Hinkel, Jacques Kaiser, Igor Peric,
  Oliver Denninger, et~al.
\newblock Connecting artificial brains to robots in a comprehensive simulation
  framework: the neurorobotics platform.
\newblock {\em Frontiers in neurorobotics}, 11:2, 2017.

\bibitem{markram2012human}
Henry Markram.
\newblock The human brain project.
\newblock {\em Scientific American}, 306(6):50--55, 2012.

\bibitem{mujocopy}
Alex Ray, Bob McGrew, Jonas Schneider, Jonathan Ho, Peter Welinder, Wojciech
  Zaremba, and Jerry Tworek.
\newblock mujoco-py.
\newblock \url{https://github.com/openai/mujoco-py}, 2020.

\bibitem{slotine1988robust}
Jean-Jacques~E Slotine, Oussama Khatib, and D~Ruth.
\newblock Robust control in operational space for goal-positioned manipulator
  tasks.
\newblock {\em International Journal of Robotics \& Automation}, 3(1):28--34,
  1988.

\bibitem{dewolf2016spiking}
Travis DeWolf, Terrence~C Stewart, Jean-Jacques Slotine, and Chris Eliasmith.
\newblock A spiking neural model of adaptive arm control.
\newblock {\em Proceedings of the Royal Society B: Biological Sciences},
  283(1843):20162134, 2016.

\bibitem{macneil2011fine}
David MacNeil and Chris Eliasmith.
\newblock Fine-tuning and the stability of recurrent neural networks.
\newblock {\em PloS one}, 6(9), 2011.

\bibitem{rasmussen2019nengodl}
Daniel Rasmussen.
\newblock Nengodl: Combining deep learning and neuromorphic modelling methods.
\newblock {\em Neuroinformatics}, 17(4):611--628, 2019.

\bibitem{morcos2019nengofpga}
Benjamin Morcos.
\newblock Nengofpga: an fpga backend for the nengo neural simulator.
\newblock Master's thesis, University of Waterloo, 2019.

\bibitem{nengoloihi}
Eric Hunsberger, Trevor Bekolay, Daniel Rasmussen, Aaron Voelker, Terry
  Stewart, Kinjal Patel, Travis DeWolf, and Chris Eliasmith.
\newblock Nengoloihi.
\newblock \url{https://github.com/nengo/nengo-loihi}, 2018.

\bibitem{mundy2015efficient}
Andrew Mundy, James Knight, Terrence~C Stewart, and Steve Furber.
\newblock An efficient spinnaker implementation of the neural engineering
  framework.
\newblock In {\em 2015 International Joint Conference on Neural Networks
  (IJCNN)}, pages 1--8. IEEE, 2015.

\end{thebibliography}

\end{document}